\def\year{2015}
\documentclass[letterpaper]{article}
\usepackage{aaai}
\usepackage{times}
\usepackage{helvet}
\usepackage{courier}
\usepackage{amsmath}
\usepackage{amssymb}
\usepackage{multirow}
\usepackage{color}
\usepackage{graphicx}
\usepackage[ruled,linesnumbered]{algorithm2e}
\def\argmax{\operatornamewithlimits{arg\,max}}
\def\w{{\mathbf w}}
\def\q{{\mathbf q}}
\def\d{{\mathbf d}}

\def\y{{\mathbf y}}
\def\0{{\mathbf 0}}
\def\v{{\mathbf v}}
\def\a{{\boldsymbol \alpha}}
\def\b{{\boldsymbol \beta}}
\def\T{{\mathsf{T}}}
\def\C{{C}} 
\def\H{{H}} 
\def\J{{J}}
\def\M{{\mathbf M}}
\def\D{{\mathbf D}}
\def\Fai{{\mathbf \Phi}}
\def\fai{{\boldsymbol \phi}}
\def\Y{{\mathcal Y}}
\def\R{{\mathbb R}}

\def\A{{\mathcal A}}
\def\l21{{\ell_{2,1}}}
\def\best{\textbf}
\def\second{}

\begin{document}
\title{Metric Learning Driven Multi-Task Structured Output Optimization \\for Robust Keypoint Tracking}
\author{Liming Zhao, Xi Li\thanks{Corresponding Author}, Jun Xiao, Fei Wu, Yueting Zhuang\\
College of Computer Science\\
Zhejiang University, Hangzhou 310027, China\\
\{zhaoliming, xilizju, junx, wufei, yzhuang\}@zju.edu.cn\\
}
\maketitle

\begin{abstract}
\begin{quote}
As an important and challenging problem in computer vision and graphics, keypoint-based object tracking is typically formulated in a spatio-temporal statistical learning framework. However, most existing keypoint trackers are incapable of effectively modeling and balancing the following three aspects in a simultaneous manner:  temporal model coherence across frames, spatial model consistency within frames, and discriminative feature construction. To address this issue, we propose a robust keypoint tracker based on spatio-temporal multi-task structured output optimization driven by discriminative metric learning. Consequently, temporal model coherence is characterized by multi-task structured keypoint model learning over several adjacent frames, while spatial model consistency is modeled by solving a geometric verification based structured learning problem. Discriminative feature construction is enabled by metric learning to ensure the intra-class compactness and inter-class separability. Finally, the above three modules are simultaneously optimized in a joint learning scheme. Experimental results have demonstrated the effectiveness of our tracker.
\end{quote}
\end{abstract}

\section{Introduction}
Due to the effectiveness and efficiency in object motion analysis, keypoint-based object tracking~\cite{KLT1981,Optical2010,keypoint2013,ConKey} is a popular and powerful tool of video processing, and thus has a wide range of applications such as augmented reality (AR), object retrieval, and video compression. By encoding the local structural information on object appearance~\cite{Li2013}, it is generally robust to various appearance changes caused by several complicated factors such as shape deformation, illumination variation, and partial occlusion~\cite{EvaDes,StrKey}. Motivated by this observation, we focus on constructing effective and robust keypoint models to well model the intrinsic spatio-temporal structural properties of object appearance in this paper.

\begin{figure}[t]
\centering
\begin{minipage}[t]{1.0\linewidth}
    \centering
    \includegraphics[width=1\textwidth]{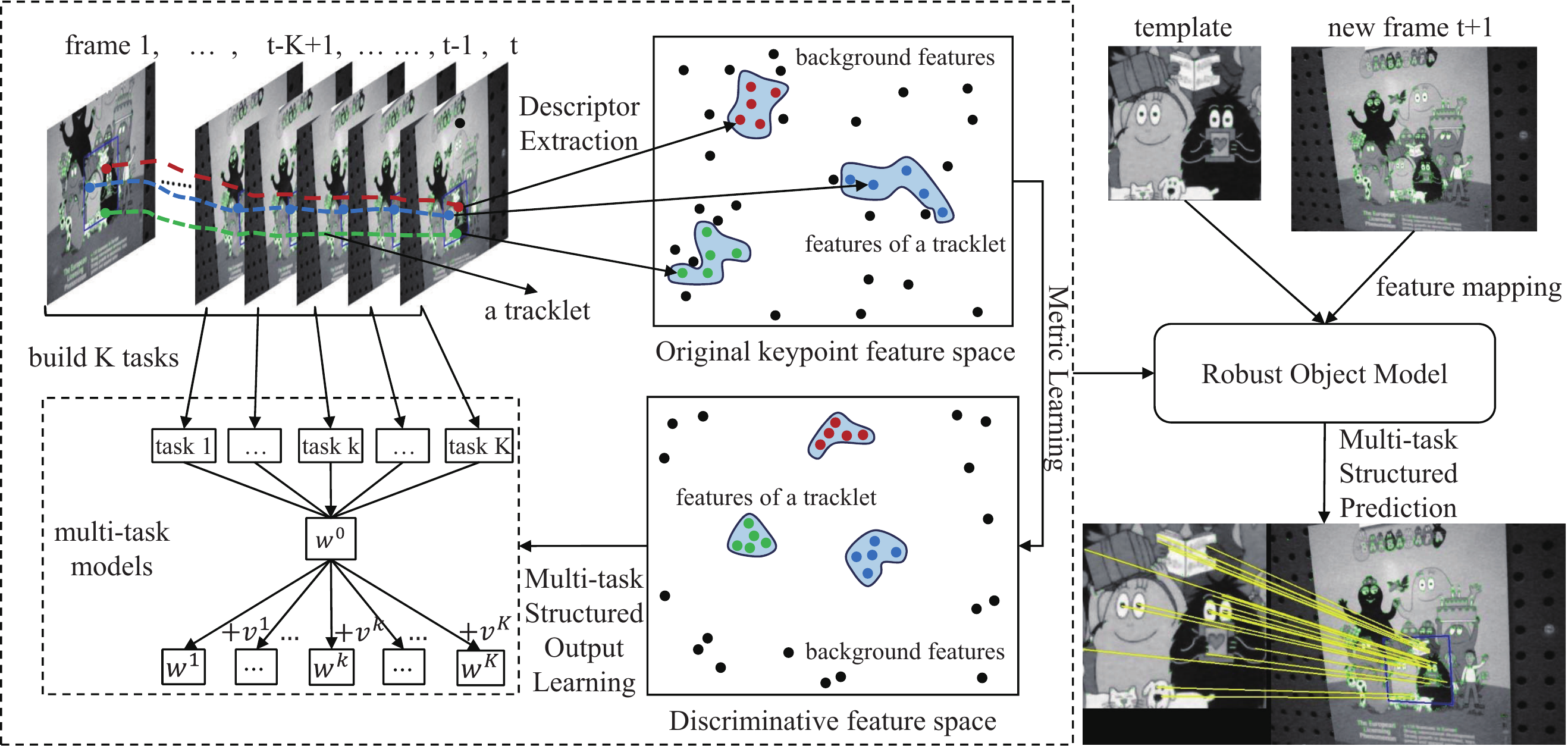}
\end{minipage}
\caption{\footnotesize{Illustration of our tracking approach.}}
\label{fig:firstShow}
\end{figure}

Typically, keypoint model construction consists of keypoint representation and statistical modeling. For keypoint representation, a variety of keypoint descriptors are proposed to encode the local invariance information on object appearance, for example, SIFT~\cite{SIFT} and SURF~\cite{SURF}. To further speed up the feature extraction process, a number of binary local descriptors emerge, including BRIEF~\cite{Brief}, ORB~\cite{ORB}, BRISK~\cite{BRISK}, FREAK~\cite{FREAK}, etc. Since the way of feature extraction is handcrafted and fixed all the time, these keypoint descriptors are usually incapable of effectively and flexibly adapting to complex time-varying appearance variations as tracking proceeds.

In general, statistical modeling is cast as a tracking-by-detection problem, which seeks to build an object locator based on discriminative learning such as randomized decision trees~\cite{RandTrees,RandFerns} and boosting~\cite{Boost2007,Boost2013}. However, these approaches usually generate the binary classification output for object tracking, and thus ignore the intrinsic structural or geometrical information (e.g., geometric transform across frames) on object localization and matching during model learning. To address this issue, Hare {\it et al.}~\cite{Hare2012} propose a structured SVM-based keypoint tracking approach that incorporates the RANSAC-based geometric matching information into the optimization process of learning keypoint-specific SVM models. As a result, the proposed tracking approach is able to simultaneously find correct keypoint correspondences and estimate underlying object geometric transforms across frames. In addition, the model learning process is independently carried out frame by frame, and hence ignores the intrinsic cross-frame interaction information on temporal model coherence, leading to instable tracking results in complicated scenarios.

In this work, we propose a joint learning approach that is capable of well balancing the following three important parts: temporal model coherence across frames, spatial model consistency within frames, and discriminative feature construction. As illustrated in Figure~\ref{fig:firstShow}, the joint learning approach ensures the temporal model coherence by building a multi-task structured model learning scheme, which encodes the cross-frame interaction information by simultaneously optimizing a set of mutually correlated learning subtasks (i.e., a common model plus different biases) over several successive frames. As a result, the interaction information induced by multi-task learning can guide the tracker to produce stable tracking results. Moreover, the proposed approach explores the keypoint-specific structural information on spatial model consistency by performing geometric verification based structured output learning, which aims to estimate a geometric transformation while associating cross-frame keypoints. In order to make the keypoint descriptors well adapt to time-varying tracking situations, the proposed approach naturally embeds metric learning to the structured SVM learning process, which enhances the discriminative power of inter-class separability.

In summary, we propose a keypoint tracking approach that learns an effective and robust keypoint model through metric learning-driven multi-task structured output optimization. The main contributions of this work are as follows:
\begin{enumerate}
\item
We propose a multi-task joint learning scheme to learn structured keypoint models by simultaneously considering spatial model consistency, temporal model coherence, and discriminative feature learning. An online optimization algorithm is further presented to efficiently and effectively solve the proposed scheme. To our knowledge, it is the first time that such a joint learning scheme is proposed for learning-based keypoint tracking.
\item
We create and release a new benchmark video dataset containing four challenging video sequences (covering several complicated scenarios) for experimental evaluations. In these video sequences, the keypoint tracking results are manually annotated as ground truth. Besides, the quantitative results on them are also provided in the experimental section.
\end{enumerate}

\section{Approach}
Our tracking approach is mainly composed of two parts: learning part and prediction part. Namely, an object model is first learned by a multi-task structured learning scheme in a discriminative feature space (induced by metric learning). Based on the learned object model, our approach subsequently produces the tracking results through structured prediction. Using the tracking results, a set of training samples are further collected for structured learning. The above process is repeated as tracking proceeds.

\subsection{Preliminary}\label{sec:model}
Let the template image $O$ be represented as a set of keypoints $O=\{(u_i, \q_i)\}_{i=1}^{N^O}$, where each keypoint is defined by a location $u_i$ and associated descriptor $\q_i$. Similarly, let $I=\{(v_j,\d_j)\}_{j=1}^{N^I}$ denote the input frame with keypoints. Typically, the traditional approaches construct the correspondences between the template keypoints and the input frame keypoints. The correspondences are scored by calculating the distances between $\{\q_i\}_{i=1}^{N^O}$ and $\{\d_j\}_{j=1}^{N^I}$. Following the model learning approaches~\cite{Hare2012,track2012}, we learn a model parameterized by a weight vector $\w_i$ for the template keypoint $u_i$ to score each correspondence. The set of the hypothetical correspondences is defined as $\C=\{(u_i,v_j,s_{ij})| (u_i,\q_i)\in O,(v_j,\d_j)\in I,s_{ij}=\left<\w_i,\d_j\right>\}$, where $s_{ij}$ is a correspondence score and $\left<\cdot,\cdot\right>$ is the inner product.

Similar to~\cite{Hare2012,track2013}, we estimate the homography transformation for planar object tracking  as the tracking result based on the hypothetical correspondences.

\subsection{Multi-task Structured Learning}\label{sec:structure}
During the tracking process, the keypoints in the successive frames $\{I_1,I_2,\dots\}$ corresponding to the $i$-th keypoint $u_i$ in the template image form a tracklet $\{v^1,v^2,\dots\}$. Based on the observation that the adjacent keypoints in a tracklet are similar to each other, the models learned for the frames $\{\w_i^1,\w_i^2,\dots\}$ should be mutually correlated. So we construct $K$ learning tasks over several adjacent frames. For example, task $k$ learns a model $\w^k$ over the training samples collected from the frames $I_{1}$ to $I_{t+k}$, where $\w^k=[\w_1^k,\dots,\w_{N^O}^k]^\T$ is the column concatenation of the model parameter vectors. We model each $\w^k$ as a linear combination of a common model $\w^0$ and an unique part $\v^k$~\cite{multi2013}:
\begin{equation}
\w^k=\w^0+\v^k,\quad k=1,\dots,K
\end{equation}
where all the vectors $\{\v^k\}_{k=1}^K$ are ``small'' when the tasks are similar to each other.

To consider the spatial model consistency in the model learning process, the transformation which maps the template to the location of the input frame is regarded as a structure, which can be learned in a geometric verification based structured learning framework. In our approach, the expected transformation $\hat{\y}$ is expressed as $\hat{\y}=\argmax_{\y\in\Y}F(\C,\y)$, where $F$ is a compatibility function, scoring all possible transformations generated by using the RANSAC~\cite{Ransac} method. Before introducing the compatibility function, we give the definition of the inlier set with a specific transformation $\y$:
\begin{equation}\label{eq:inlier}
\H(\C,\y)=\{(u_i,v_j)| (u_i,v_j)\in\C, \lVert\y(u_i)-v_j\rVert<\tau\}
\end{equation}
where $\y(u_i)$ is the transformed location in the input frame of the template keypoint location $u_i$, $\tau\in\R$ is a spatial distance threshold, and $\lVert\cdot\rVert$ denotes the Euclidean norm.

The compatibility function with respect to task $k$ is then defined as the total score of the inliers:
\begin{equation}\label{eq:F}
\begin{aligned}
F^k(\C,\y)=\hspace{-1em}\sum_{(u_i,v_j)\in\H(\C,\y)}\hspace{-2em}\left<\w_i^k,\d_j\right>
        =\left<\w^k,\Fai(\C,\y)\right>
\end{aligned}
\end{equation}
where $\Fai(\C,\y)$ is a joint feature mapping vector concatenated by $\fai_i(\C,\y)$ which is defined as:
\begin{equation}\label{eq:Phi}
\fai_i(\C,\y)=
\begin{cases}
\d_j & \exists(u_i,v_j)\in \C: \lVert\y(u_i)-v_j\rVert<\tau\\
\0   &\textrm{otherwise}
\end{cases}
\end{equation}

Given training samples $\{(\C_t,\y_t)\}_{t=1}^{T}$ (each $\C_t$ is the hypothetical correspondences of the frame $I_t$, and $\y_t$ is the predicted transformation), a structured output maximum margin framework~\cite{markov,SSVM2005} is used to learn all the multi-task models, which can be expressed by the following optimization problem:
\begin{equation}\label{eq:multi}
\begin{aligned}
\min_{\w^0,\v^k,\boldsymbol\xi}
&\frac{1}{2}\lVert\w^0\rVert^2
+\frac{\lambda_1}{2K}\sum_{k=1}^K\lVert\v^k\rVert^2
+\nu_1\sum_{k=1}^K\sum_{t=k}^T\xi_{kt}
\\
\text{s.t.}
&\forall k,t, \xi_{kt} \geq 0\\
&\forall k,t, \forall \y\neq\y_t: \delta F_t^k(\y)\geq\Delta(\y_t,\y)-\xi_{kt}
\end{aligned}
\end{equation}
where $\delta F^k_t(\y)=\left<\w^k,\Fai(\C_t,\y_t)\right>-\left<\w^k,\Fai(\C_t,\y)\right>$ and $\Delta(\y_t,\y)$ is a loss function which measures the difference of two transformations (in our case, the loss function $\Delta(\y_t,\y)=|\#\H(\C,\y_t)-\#\H(\C,\y)|$ is the difference in number of two inlier sets). The nonnegative ${\lambda_1}$ is the weight parameter for multiple tasks, and the weighting parameter $\nu_1$ determines the trade-off between accuracy and regularization.

\begin{figure}[t]\footnotesize
\centering
\begin{minipage}{0.49\linewidth}
    \centering
    \includegraphics[width=1\textwidth]{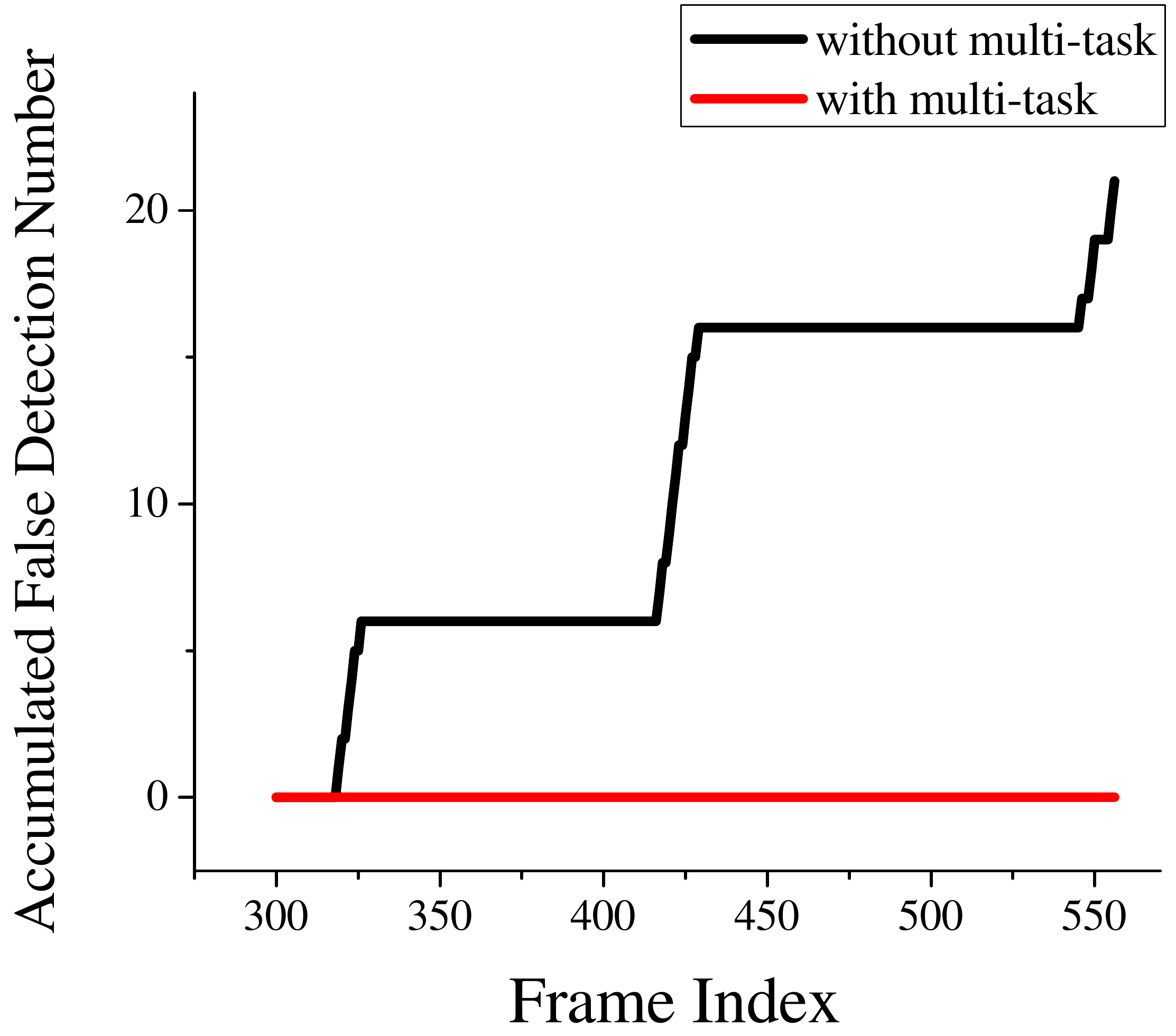} \\
    (a) tracking result
\end{minipage}
\begin{minipage}{0.49\linewidth}
    \centering
    \includegraphics[width=0.82\textwidth]{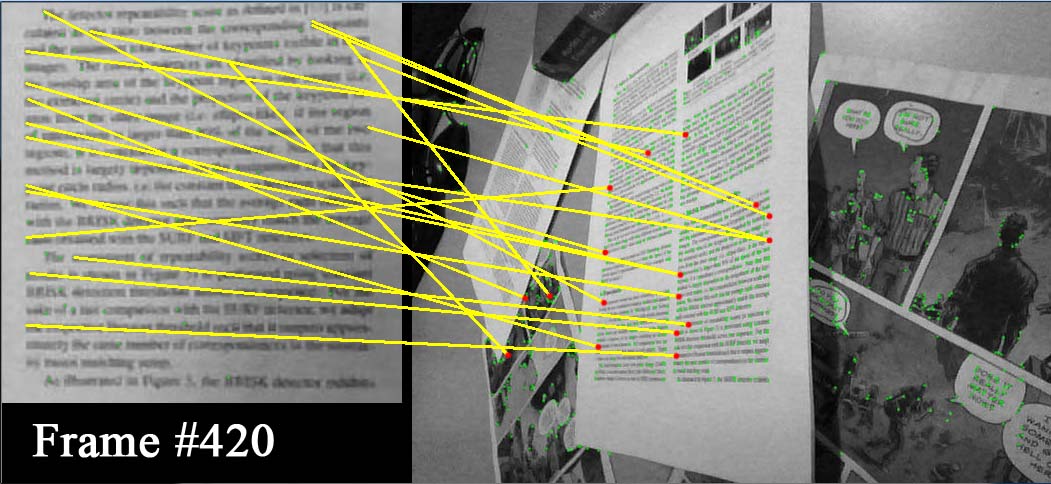} \\
    (b) without multi-task
    \includegraphics[width=0.82\textwidth]{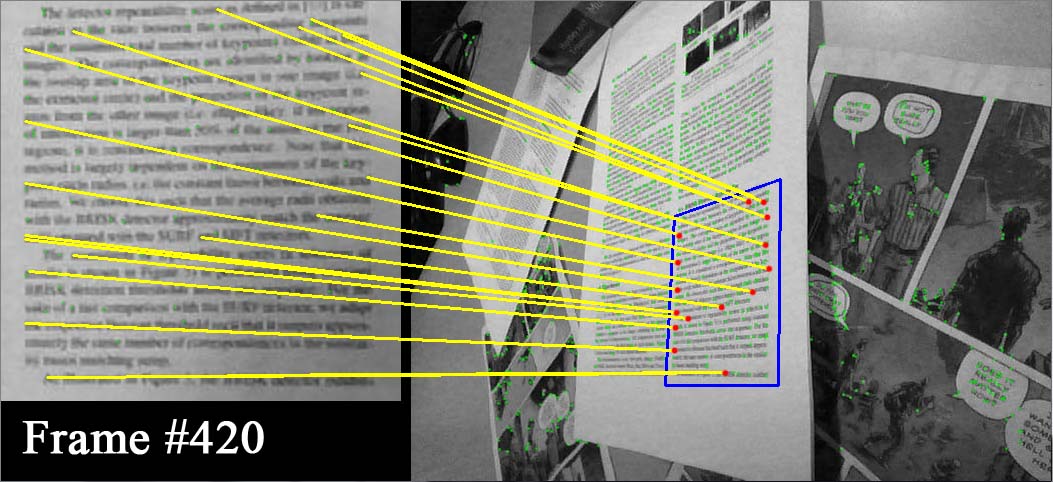} \\
    (c) with multi-task
\end{minipage}
\caption{\footnotesize{Example tracking results. Figure (a) shows the quantitative results of the trackers with and without multi-task learning in the accumulated number of falsely detected frames (lower is better), the tracker with multi-task learning produces a stable tracking result. Figure (b) and (c) show the qualitative tracking results. The blue bounding box represents the location of the detected object ,and the yellow line represents a keypoint correspondence. In figure (b), the tracker without multi-task learning fails to match keypoints correctly.}}
\label{fig:multi}
\end{figure}

To better describe the contribution of the multi-task learning, example tracking results of the trackers with and without multi-task learning are shown in Figure~\ref{fig:multi}. From Figure~\ref{fig:multi}(b), we observe that the independent model fails to match the keypoints in the case of drastic rotations, while the multi-task model enables the temporal model coherence to capture the information of rotational changes, thus produces a stable tracking result.

\subsection{Discriminative Feature Space}\label{sec:metric}
In order to make the keypoint descriptors well adapt to time-varying tracking situations, we wish to learn a mapping function $f(\d)$ that maps the original feature space to another discriminative feature space, in which the semantically similar keypoints are close to each other while the dissimilar keypoints are far away from each other, that can be formulated as a metric learning process~\cite{Metric2009,Metric2011}. We then use the mapped feature $f(\d)$ to replace the original feature $\d$ in the structured learning process, to enhance its discriminative power of inter-class separability.

\begin{figure}[t]\footnotesize
\centering
\begin{minipage}{0.49\linewidth}
    \centering
    \includegraphics[width=1\textwidth]{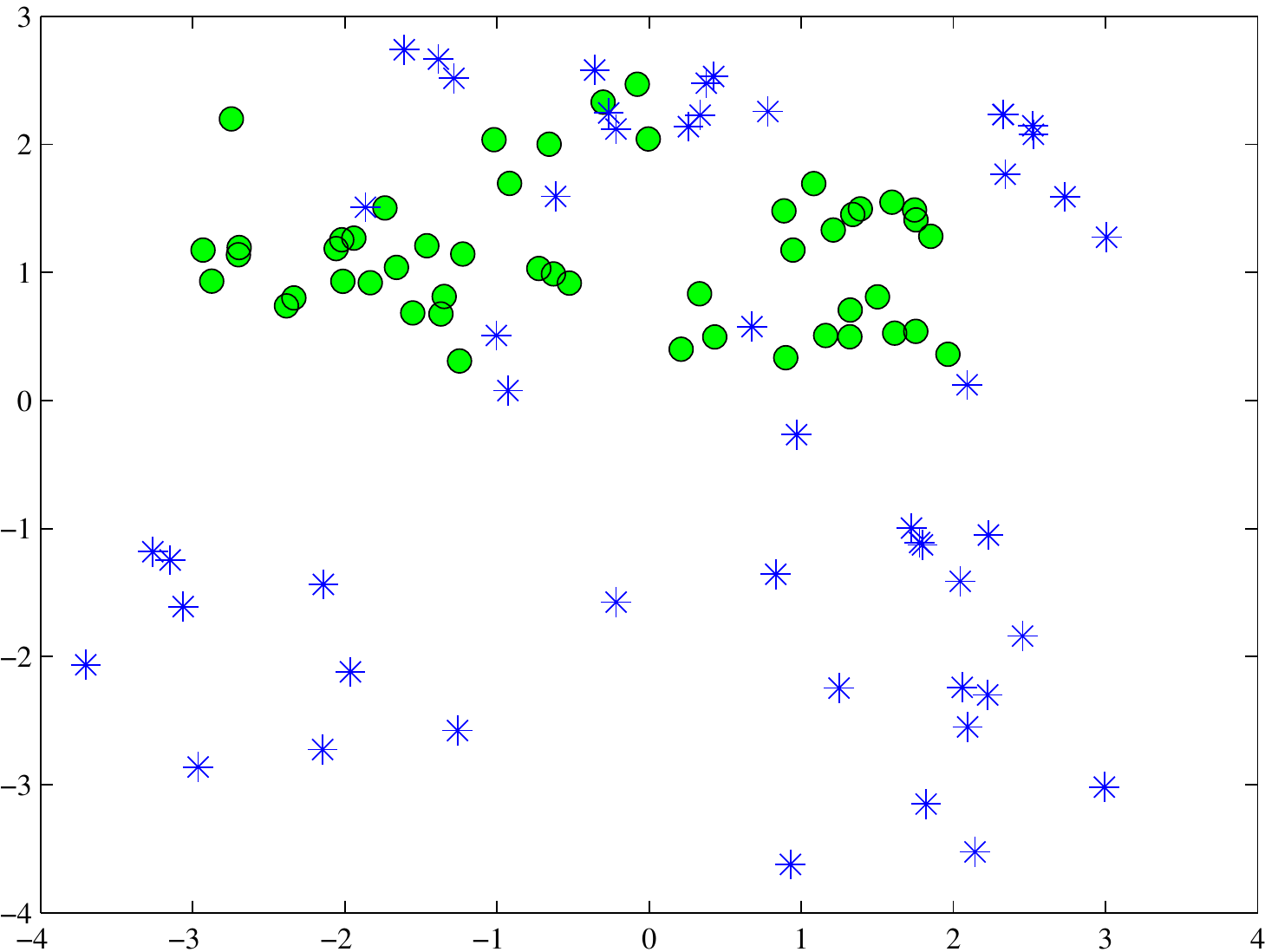}
    (a) before mapping
\end{minipage}
\begin{minipage}{0.49\linewidth}
    \centering
    \includegraphics[width=1\textwidth]{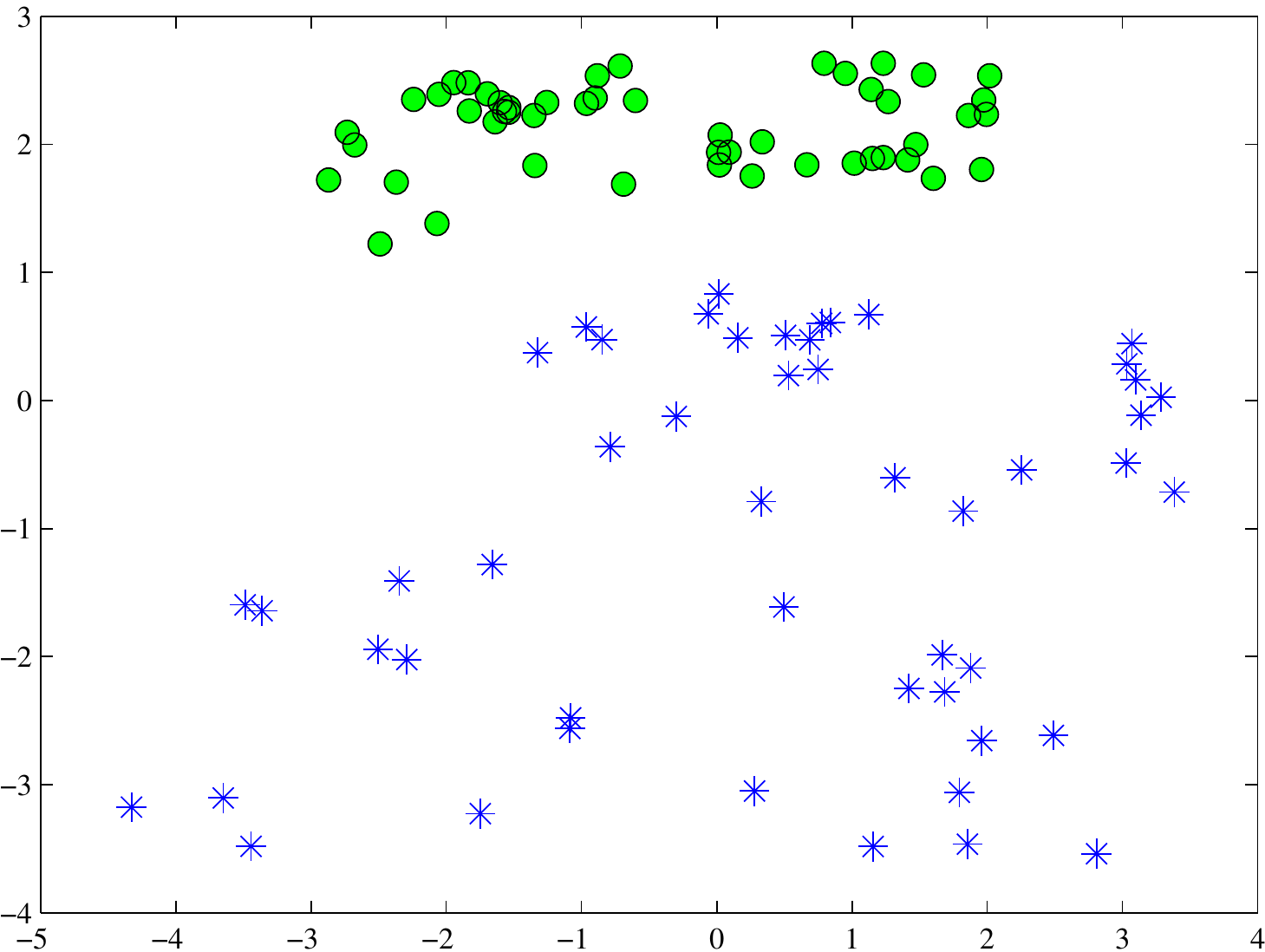}
    (b) after mapping
\end{minipage}
\caption{\footnotesize{Visualization of keypoint features using PCA. The $50$ green circle points represent the keypoints from $50$ successive frames corresponding to the same keypoint in the template (semantically similar keypoints). The blue asterisk points represent any other keypoints, which are dissimilar to the above $50$ keypoints. In figure (a), all the keypoints mix together in the original feature space. In figure (b), there is a large margin between dissimilar keypoints in current learned feature space.}}
\label{fig:metric}
\end{figure}

Figure~\ref{fig:metric} shows an example of such feature space transformation. Before the mapping procedure, the object keypoints and the background keypoints can not be discriminated in the original feature space. After the transformation, the keypoints in different frames corresponding to the same keypoint in the template, which are semantically similar, get close to each other in the mapped feature space, while the features of the other keypoints have a distribution in another side with a large margin.

The following describes how to learn the mapping function. For a particular task $k$, given the learned model $\w_i^k$, the distance between a doublet $(\d_j,\d_k)$ is defined as follows:
\begin{equation}\label{eq:distance}
D_i^k(\d_j,\d_{j'})=\left<\w_i^k,f(\d_j)-f(\d_{j'})\right>
\end{equation}

We assume that the binary matrix $p_{j{j'}}\in\{0,1\}$ indicates whether or not the features $\d_j$ and $\d_{j'}$ are semantically similar (if they are similar, $p_{j{j'}}=1$). Therefore, the hinge loss function on a doublet is defined as:
\begin{equation}
\ell_i^k(\d_{j},\d_{j'})=[(-1)^{p_{j{j'}}}(1-D_i^k(\d_{j},\d_{j'}))]_+
\end{equation}
where $[z]_+=\max(z,0)$.

To learn the effective feature consistently in our mapping process, we wish to find the group-sparsity of the features. So we utilize $\l21$-norm~\cite{L21norm1,L21norm2} to learn the discriminative information and feature correlation consistently. Since we use a linear transformation $f(\d)=\M^{\T}\d$ as our mapping function, the $\l21$-norm for the mapping matrix $\M$ is defined as: $\lVert\M\rVert_{2,1}=\sum_{i}\sqrt{\sum_{j}\M_{ij}^2}$.

Given all the keypoint features from the video frames $\{I_t\}_{t=1}^T$, we collect all possible combinations of the features as the training set, which is denoted as $\A=\{(\d_j,\d_{j'})|\d_j\in\{I_t\}_{t=1}^T, {j'}\neq{j}, \d_{j'}\in\{I_t\}_{t=1}^T\}$. We obtain the binary matrix $p_{j{j'}}$ by using the tracking results (if $\d_j$ and $\d_{j'}$ from different frames correspond to the same keypoint in the template, $p_{j{j'}}$ is set to 1; otherwise, $p_{j{j'}}$ is set to 0). We wish to minimize the following cost function consisting of the empirical loss term and the $\l21$-norm regularization term:
\begin{equation}\label{eq:metricLoss}
\sum_{i,k,(\d_{j},\d_{j'})\in\A}\hspace{-1.2em}\ell_i^k(\d_{j},\d_{j'})
 + \lambda\lVert\M\rVert_{2,1}
\end{equation}

The cost function is incorporated into our multi-task structured learning framework, and then a unified joint learning scheme for object tracking is obtained. The final optimization problem of our approach is expressed in the following form:
\begin{equation}\label{eq:final}
\begin{aligned}
\min_{\w^0,\v^k,\M,\boldsymbol\xi,\boldsymbol\gamma}
&\frac{1}{2}\lVert\w^0\rVert^2
+\frac{\lambda_1}{2K}\sum_{k=1}^K\lVert\v^k\rVert^2
+{\lambda_2}\lVert\M\rVert_{2,1}&\\
+\sum_{k=1}^K&
\Big(
\nu_1\sum_{t=k}^T\xi_{kt}
+\nu_2\sum_{t=k}^T\sum_{(u_i,v_j) \in \H(\C_t,\y_t)}\hspace{-1.5em}\gamma_{kti}
\hspace{0.3em}\Big)\\
\text{s.t.}\forall k,t,& \xi_{kt} \geq 0&\\
\forall k,t,& \forall \y\neq\y_t: \delta F_t^k(\y)\geq\Delta(\y_t,\y)-\xi_{kt}&\\
\forall k,t,&i: \gamma_{kti}\geq 0&\\
\forall k,t,&(u_i,v_j), \forall j'\neq j: D_i^k(\d_j,\d_{j'})\geq 1-\gamma_{kti}&
\end{aligned}
\end{equation}

After all the models $\w^1,\w^2,\dots,\w^K$ are learned, we use the last model $\w=\w^K$ to predict the result of new frame $I_t$. We use the RANSAC method to generate hypothetical transformations. Based on the model $\w$, we predict the expected transformation $\y_t$ from all hypothetical transformations by maximizing Eq.~\eqref{eq:F}. The hypothetical correspondence set $\C_t$ of the frame $I_t$ and the predicted transformation $\y_t$ are then added to our training set. We use all the training samples collected from the results of previous $K$ frames ($I_{t-K+1}$ to $I_{t}$) to update our model. Then the above process is repeated as tracking proceeds.

\subsection{Online Optimization}\label{sec:optimization}
The optimization problem presented in Eq.~\eqref{eq:final} can be solved online effectively. We adopt an alternating optimization algorithm to solve the optimization problem.

\vspace{-1em}
\paragraph{Unconstrained form}
Let $\a_{kt}=[\max_{\y\neq\y_t}\{\Delta(\y_t,\y)-\delta F_t^k(\y)\}]_+$
and $\b_{kti}=[\max_{j'\neq j}\{1-D_i^k(\d_j,\d_{j'})\}]_+$. Therefore, Eq.~\eqref{eq:final} can be rewritten to an unconstrained form:
\begin{equation}\label{eq:relaxed}
\begin{aligned}
\min_{\w^0,\v^k,\M}
\frac{1}{2}\lVert\w^0\rVert^2
+\frac{\lambda_1}{2K}\sum_{k=1}^K\lVert\v^k\rVert^2
+{\lambda_2}\lVert\M\rVert_{2,1}\\
+\sum_{k=1}^K
\Big(
\nu_1\sum_{t=k}^T\a_{kt}
+\nu_2\sum_{t=k}^T\sum_{(u_i,v_j) \in \H(\C_t,{\y_t})}\hspace{-1.5em}\b_{kti}
\hspace{0.2em}\Big)
\end{aligned}
\end{equation}

For descriptive convenience, let $\J$ denote the term of $\nu_1\sum_{t=k}^T\a_{kt}+\nu_2\sum_{t=k}^T\sum_{(u_i,v_j) \in \H(\C_t,{\y_t})}\b_{kti}$.

\vspace{-1em}
\paragraph{Fix $\{\v^k\}_{k=1}^K$ and $\w^0$, solve $\M$}
Firstly, we fix all $\{\v^k\}_{k=1}^K$ and $\w^0$, and learn the transformation matrix $\M$ by solving the following problem:
\begin{equation}\label{eq:M}
\min_{\M}
\hspace{0.2em}
\lVert\M\rVert_{2,1}
+\frac{1}{\lambda_2}\sum_{k=1}^K{\J}
\end{equation}

Let $\M^i$ denote the $i$-th row of $\M$, and $Tr(\cdot)$ denote the trace operator. In mathematics, the Eq.~\eqref{eq:M} can be converted to the following form:
\begin{equation}\label{eq:M2}
\begin{aligned}
\min_{\M}
\hspace{0.2em}
Tr(\M^{\T}\D\M)
+\frac{1}{\lambda_2}\sum_{k=1}^K{\J}
\end{aligned}
\end{equation}
where $\D$ is the diagonal matrix of $\M$, and each diagonal element is  $\D_{ii}=\frac{1}{2\lVert\M^i\rVert_2}$. We use an alternating algorithm to calculate $\D$ and $\M$ respectively. We calculate $\M$ with the current $\D$ by using gradient descent method, and then update $\D$ according to the current $\M$. The details of solving Eq.~\eqref{eq:M2} are shown in the supplementary file.

\vspace{-1em}
\paragraph{Fix $\M$ and $\{\v^k\}_{k=1}^K$, solve $\w^0$}
Secondly, after $\M$ is learned, let $\{\v^k\}_{k=1}^K$ have been the optimal solution of Eq.~\eqref{eq:relaxed}. Then $\w^0$ can be obtained by the combination of $\v^k$ according to~\cite{Multitask}:
\begin{equation}
\w^0=\hspace{0.2em}\frac{\lambda_1}{K}\sum_{k=1}^{K}\v^k
\end{equation}
The proof can be found in our supplementary material.

\vspace{-1em}
\paragraph{Fix $\M$ and $\w^0$, solve $\{\v^k\}_{k=1}^K$}
Finally, $\{\v^k\}_{k=1}^K$ can be learned one by one using gradient descent method. In fact, we learn $\w^k=\w^0+\v^k$ instead of $\v^k$ for convenience. Let $\bar{\w}=\frac{1}{K}\sum_{k=1}^{K}\w^k$ be the average vector of all $\w^k$. Then the optimization problem for each $\w^k$ can be rewritten as:
\begin{equation}\label{eq:v}
\begin{aligned}
\min_{\w^k}
\hspace{0.2em}
\rho_1\lVert\w^k\rVert^2
+\rho_2\lVert\w^k-\bar{\w}\rVert^2+{\J}
\end{aligned}
\end{equation}
where $\rho_1={\lambda_1}/({\lambda_1+1})$ and $\rho_2={\lambda_1^2}/({\lambda_1+1})$ (the derivation proof is given in the supplementary material).

Given training samples $\{(\C_{t-k},\y_{t-k})\}_{k=0}^{K-1}$ at time $t$, the subgradient of Eq.~\eqref{eq:v} with respect to $\w^k$ is calculated, and we perform a gradient descent step according to:
\begin{equation}\label{eq:updateV}
\begin{aligned}
\w^k\leftarrow
\hspace{0.2em}
(1-\frac{1}{t})\w^k
+\eta\rho_2\bar{\w}
-\eta\frac{\partial\J}{\partial\w^k}
\end{aligned}
\end{equation}
where $\eta={1}/{(\rho_1t+\rho_2t)}$ is the step size (the details of the term $J$ is described in the supplementary material). We repeat the procedure to obtain an optimal solution until the algorithm converges (on average converges after $5$ iterations).

All the above is summarized in Algorithm~\ref{alg:opt}, and the details are described in the supplementary material.

\begin{algorithm}[ t ]\footnotesize
\caption{\footnotesize{Online Optimization for Tracking}}
\label{alg:opt}
    \KwIn{Input frame $I_t$ and previous models $\{\w^1,\dots,\w^K\}$}
    \KwOut{The predicted transformation $\y_t$, updated models and mapping matrix for metric learning}
\tcc{The structured prediction part}
    Calculate the correspondences $\C_t$ based on the model $\w^K$\;
    Estimate hypothetical transformations $\y$ using RANSAC;
    Calculate the inlier set of each $\y$ using Eq.~\eqref{eq:inlier}\;
    Predict the expected $\y_t$ by maximizing Eq.~\eqref{eq:F}\;
\tcc{The structured learning part}
    Collect the training samples $\{(\C_{t-k},\y_{t-k})\}_{k=0}^{K-1}$\;
    \Repeat{Alternating optimization convergence}
    {
        Calculate $\J$ according to Section~\ref{sec:optimization}\;
        \For{$k=1,\dots,K$}
        {
            Update each model $\w^k$ using Eq.~\eqref{eq:updateV}\;
        }
        Update the mapping matrix $\M$ by solving Eq.~\eqref{eq:M2}\;
    }
    \Return $\y_t$, $\{\w^1,\dots,\w^K\}$ and $\M$\;
\end{algorithm}

\section{Experiments and Results}
\subsection{Experimental Settings}

\paragraph{Dataset}
The video dataset used in our experiments consists of nine video sequences. Specifically, the first five sequences are from~\cite{Hare2012}, and the last four sequences (i.e., ``chart'', ``keyboard'', ``food'', ``book'') are recorded by ourselves. All these sequences cover several complicated scenarios such as background clutter, object zooming, object rotation, illumination variation, motion blurring and partial occlusion (example frames can be found in the supplementary material).

\vspace{-1em}
\paragraph{Implementation Details}
For keypoint feature extraction, we use FAST keypoint detector~\cite{Fast} with 256-bit BRIEF descriptor~\cite{Brief}. For metric learning, the linear transformation matrix $\M$ is initialized to be an identity matrix. For multi-task learning, the number of tasks $K$ is chosen as $5$ and we update all the multi-task models frame by frame. All weighting parameters $\lambda_1,\lambda_2,\nu_1,\nu_2$ are set to $1$, and remain fixed throughout all the experiments. Similar to~\cite{Hare2012}, we consider the tracking process of estimating homography transformation on the planar object as a tracking-by-detection task.

\vspace{-1em}
\paragraph{Evaluation Criteria}
We use the same criteria as~\cite{Hare2012} with a scoring function between the predicted homography $\y$ and the ground-truth homography $\y^*$:
\begin{equation}
S(\y,\y^*)=\frac{1}{4}\sum_{i=1}^{4}\lVert\y(c_i)-\y^*(c_i)\rVert_2
\end{equation}
where $\{c_i\}_{i=1}^4=\{(-1,-1)^\T,(1,-1)^\T,(-1,1)^\T,(1,1)^\T\}$ is a normalized square. For each frame, it is regarded as a successfully detected frame if $S(\y,\y^*)<10$, and a falsely detected frame otherwise. The average success rate is defined as the number of successfully detected frames divided by the length of the sequence, which is used to evaluate the performance of the tracker. To provide the tracking result frame by frame, we present a criterion of the accumulated false detection number, which is defined as the accumulated number of falsely detected frames as tracking proceeds.

\subsection{Experimental Results}

\paragraph{Comparison with State-of-the-art Methods}
We compare our approach with some state-of-the-art approaches, including boosting based approach~\cite{Boost2007}, structured SVM (SSVM) approach~\cite{Hare2012} and a baseline static tracking approach (without model updating). All these approaches are implemented by making use of their publicly available code. We also implement our approach in C++ and OPENCV. On average, our algorithm takes 0.0746 second to process one frame with a quad-core 2.4GHz Intel Xeon E5-2609 CPU and 16GB memory. Table~\ref{tab:others} shows the experimental results of all four approaches in the average success rate. As shown in this table, our approach performs best on all sequences.

\begin{table}[t]\footnotesize
  \centering
    \begin{tabular}{|c|c|c|c|c|}\hline
    \multirow{2}{*}{Sequence} & \multicolumn{4}{c|}{Average Success Rate(\%)} \\\cline{2-5}
                & Static  & Boosting & SSVM             & Ours\\\hline
    barbapapa   & 19.7138 & 89.0302 & \second{94.1176} & \best{94.4356}\\
    comic       & 42.5000 & 57.6042 & \second{98.1250} & \best{98.8542} \\
    map         & 81.1295 & 82.0937 & \best{98.7603} & \best{98.7603}\\
    paper       & 05.0267 & 03.8502 & \second{82.7807} & \best{88.2353} \\
    phone       & 88.1491 & 84.9534 & \second{96.6711} & \best{98.4021} \\
    chart       & 13.1461 & 01.9101 & \second{53.0337} & \best{77.5281}\\
    keyboard    & 27.8607 & 57.7114 & \second{62.3549} & \best{94.5274}\\
    food        & 32.8173 & 67.4923 & \second{85.7585} & \best{99.6904}\\
    book        & 08.5616 & 08.9041 & \second{55.8219} & \best{81.6781}\\\hline
    \end{tabular}
    \caption{\footnotesize{Comparison with state-of-the-art approaches in the average success rate (higher is better). The best result on each sequence is shown in bold font. We observe that our approach performs best on all the sequences.}}
    \label{tab:others}
\end{table}

\begin{figure}[t]
\centering
\begin{minipage}[t]{0.49\linewidth}
    \centering
    \includegraphics[width=1\textwidth]{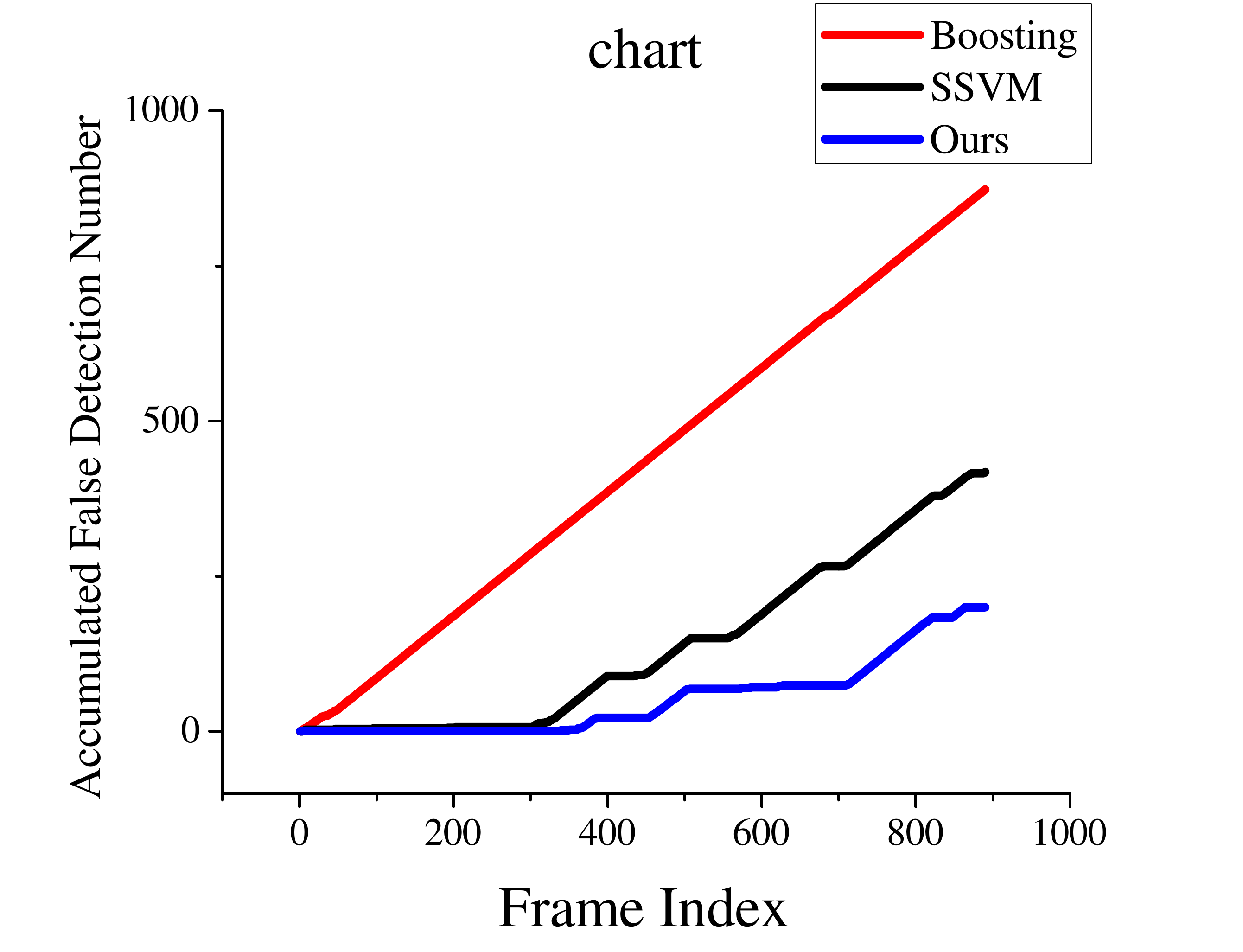}
\end{minipage}
\begin{minipage}[t]{0.49\linewidth}
    \centering
    \includegraphics[width=1\textwidth]{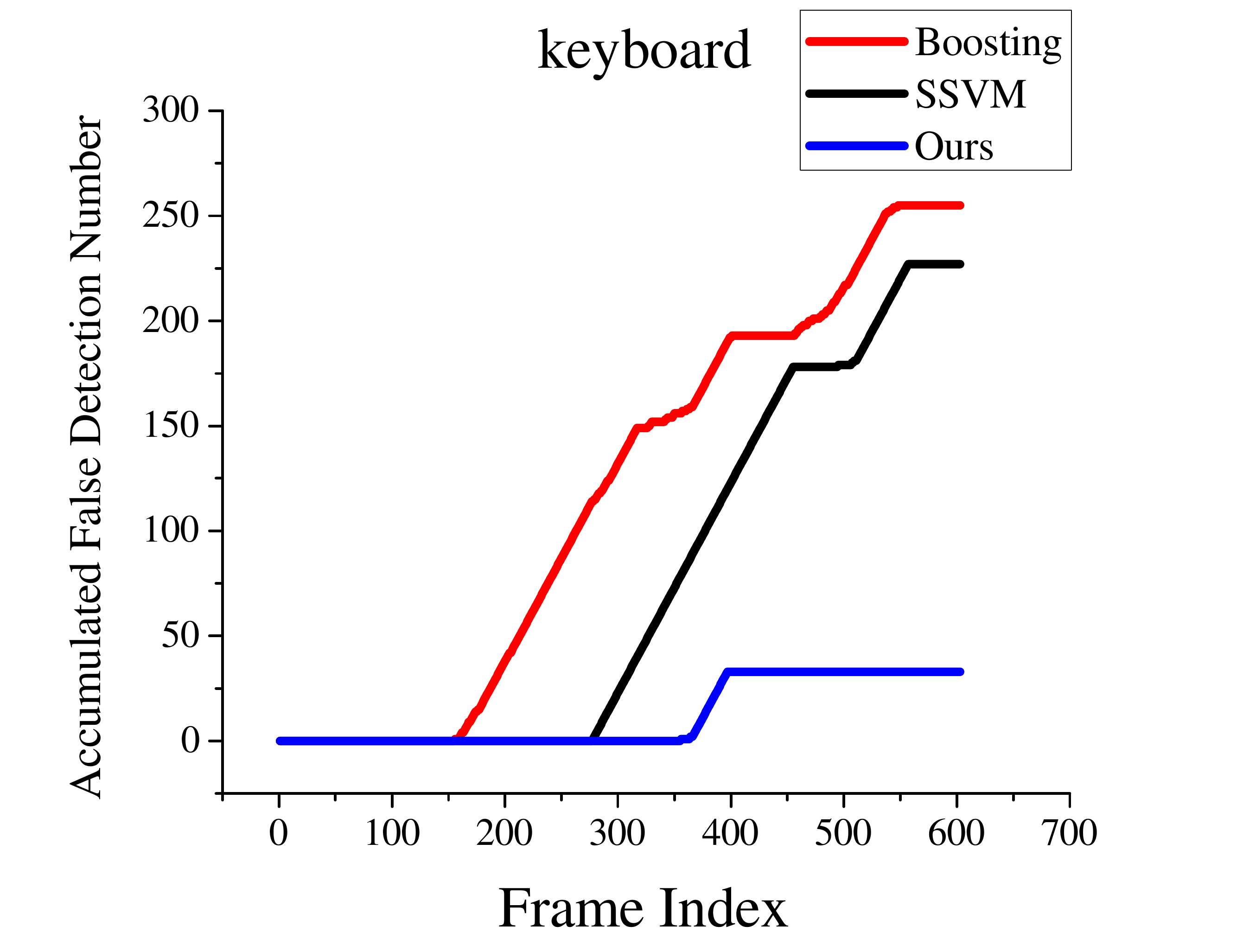}
\end{minipage}
\\
\begin{minipage}[t]{0.49\linewidth}
    \centering
    \includegraphics[width=1\textwidth]{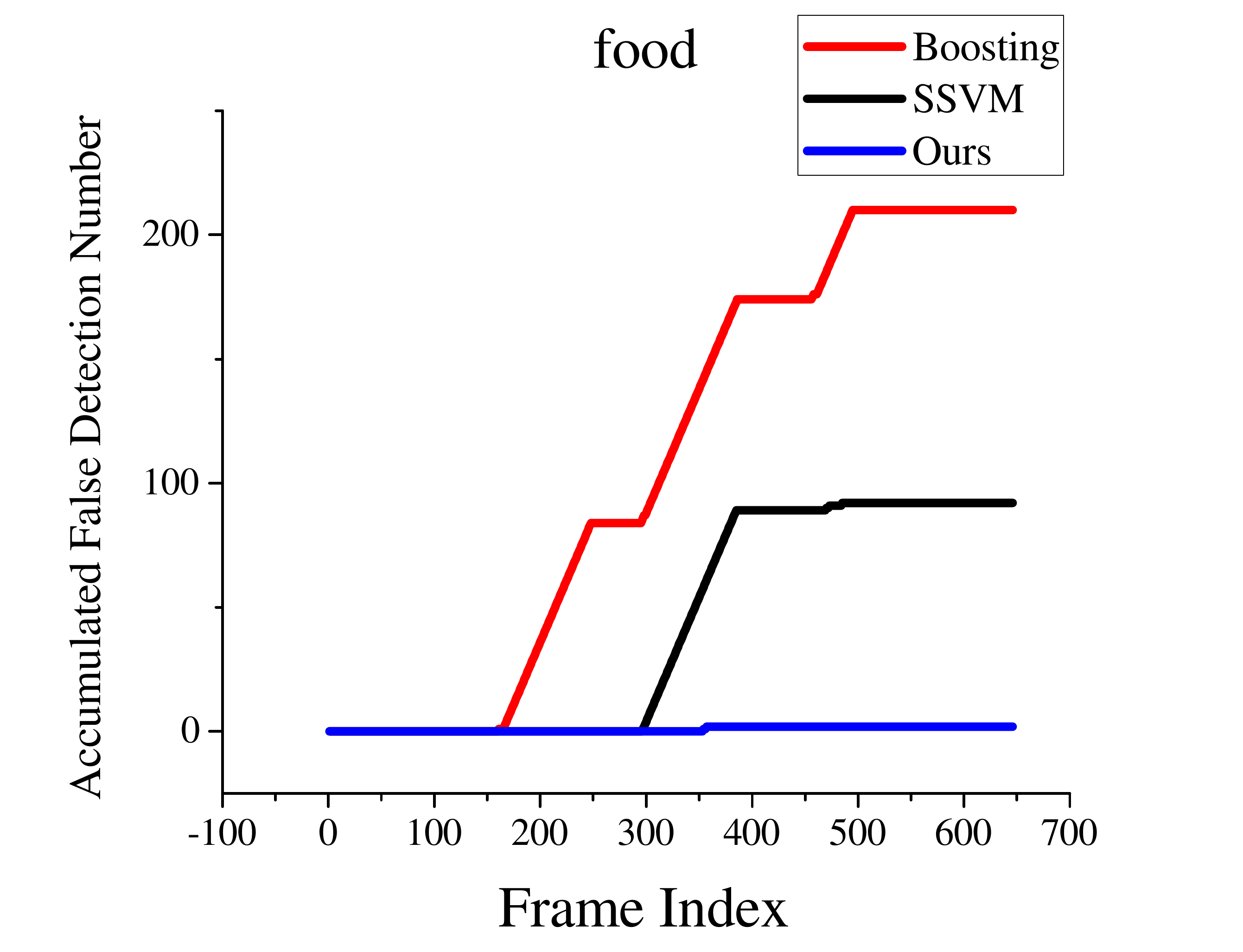}
\end{minipage}
\begin{minipage}[t]{0.49\linewidth}
    \centering
    \includegraphics[width=1\textwidth]{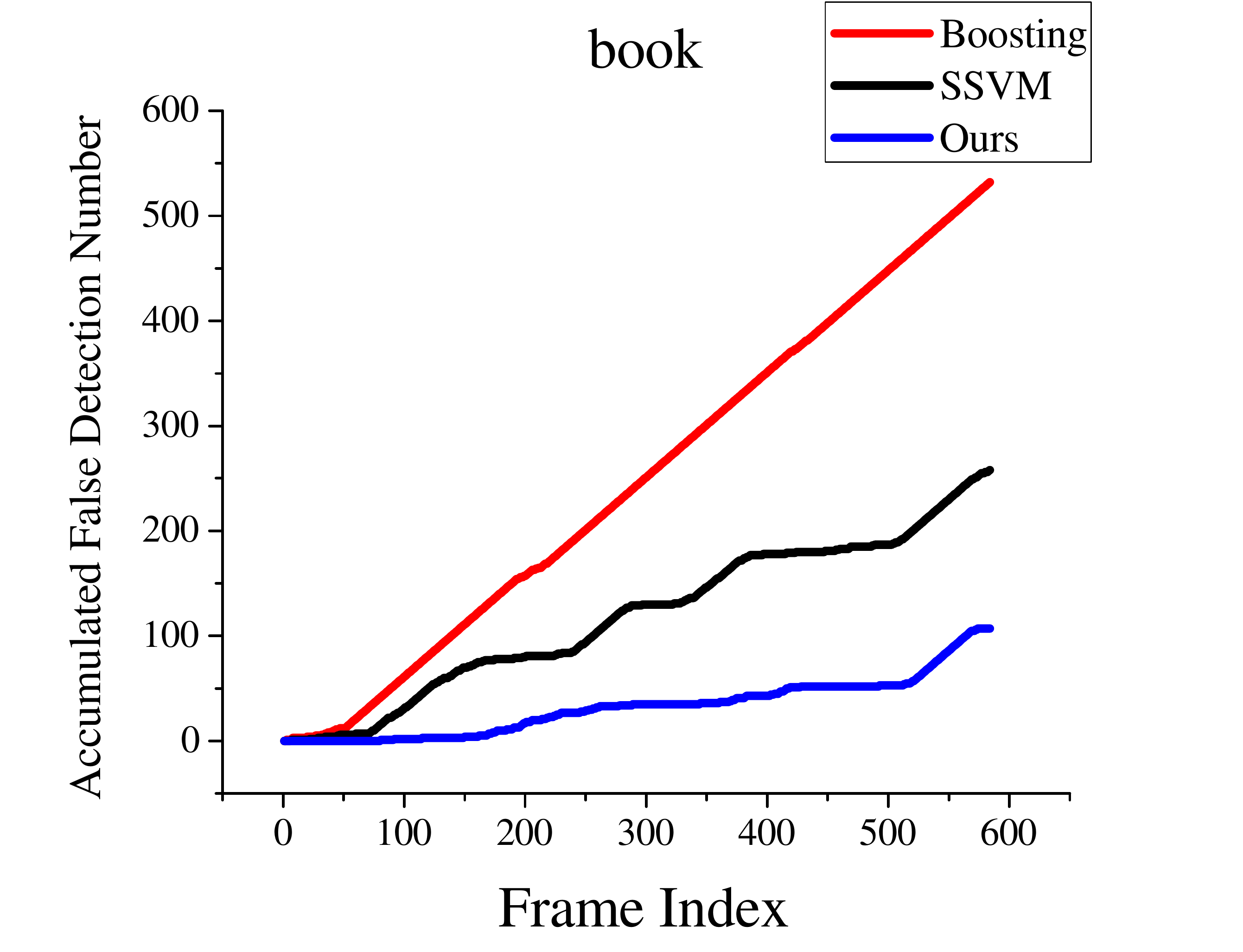}
\end{minipage}

\caption{\footnotesize{Comparison of three approaches in the accumulated number of falsely detected frames (lower is better). The curve corresponding to our approach grows slowly and is almost horizontal, which means that our tracking result is stable.}}
\label{fig:chart}
\end{figure}

To provide an intuitive illustration, we report the detection result on each frame in Figure~\ref{fig:chart}. We observe that both the ``Boosting'' and ``SSVM'' approaches obtain a number of incorrect detection results on some frames of the test sequences, while our approach achieves stable tracking results in most situations (the curve corresponding to our approach grows slowly and is almost horizontal).

\begin{figure}[t]\footnotesize
\centering
\begin{minipage}[t]{1\linewidth}
    \centering
    SSVM\quad\quad\quad\quad\quad\quad\quad\quad\quad\quad Ours\\
    \includegraphics[width=0.49\textwidth]{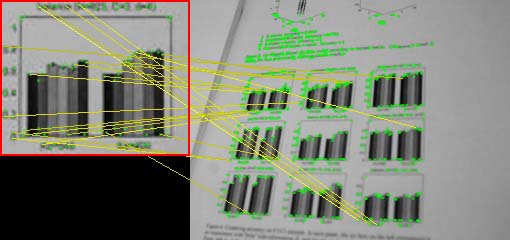}
    \includegraphics[width=0.49\textwidth]{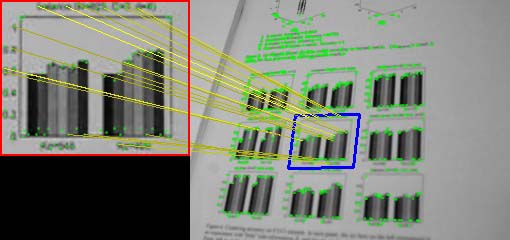} \\
    (a) chart frame 388 (camera motion blurring)\\
\end{minipage}\\
\begin{minipage}[t]{1\linewidth}
    \centering
    \includegraphics[width=0.49\textwidth]{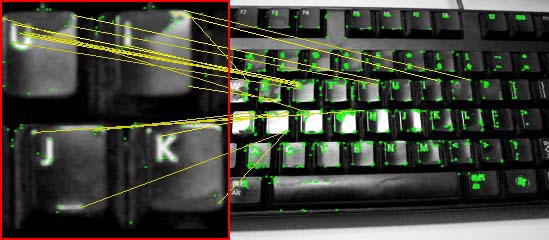}
    \includegraphics[width=0.49\textwidth]{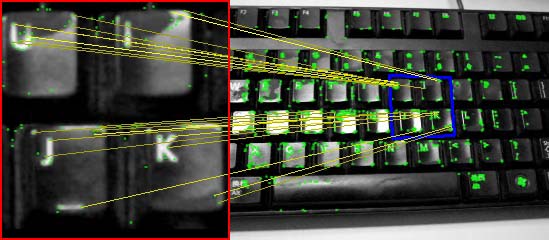} \\
    (b) keyboard frame 300 (illumination variation)\\
\end{minipage}\\
\begin{minipage}[t]{1\linewidth}
    \centering
    \includegraphics[width=0.49\textwidth]{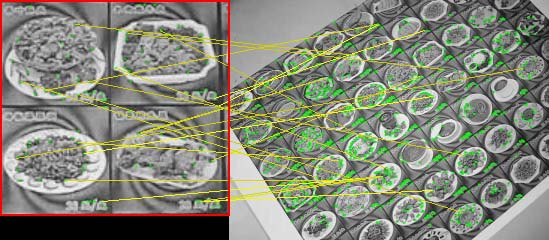}
    \includegraphics[width=0.49\textwidth]{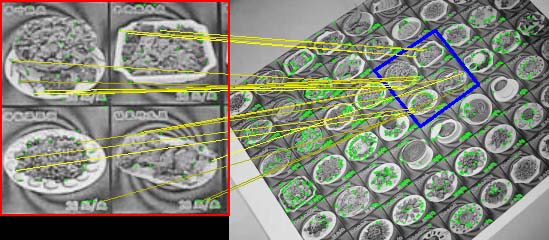} \\
    (c) food frame 350 (object rotation)\\
\end{minipage}\\
\begin{minipage}[t]{1\linewidth}
    \centering
    \includegraphics[width=0.49\textwidth]{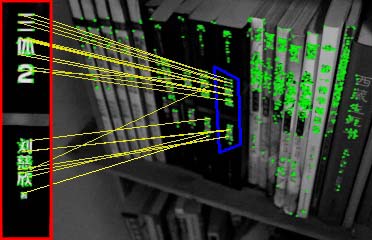}
    \includegraphics[width=0.49\textwidth]{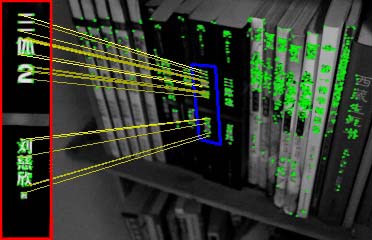} \\
    (d) book frame 357 (confusing keypoints)\\
\end{minipage}
\caption{\footnotesize{Example tracking results on our test video sequences. In each picture, the left part highlighted in red bounding box is the template image. The blue box shows the location of the detected object in the frame. Our model has adapted to obtain correct detection results in the complicated scenarios with drastic object appearance changes.}}
\label{fig:exp}
\end{figure}

Figure~\ref{fig:exp} shows the tracking results on some sample frames (more experimental results can be found in our supplementary materials). These sequences containing background clutter are challenging for keypoint based tracking. In terms of metric learning and multi-task learning, our approach still performs well in some complicated scenarios with drastic object appearance changes.

\vspace{-1em}
\paragraph{Evaluation of Our Individual Components}
To explore the contribution of each component in our approach, we compare the performances of the approaches with individual parts, including SSVM(structured SVM), SML(SSVM $+$ metric learning), SMT(SSVM $+$ multi-task learning), and SMM (SSVM $+$ ML $+$ MT, which is exactly our approach). The experimental results of all these approaches in the average success rate are reported in Table~\ref{tab:ours}.

\begin{table}[t]\footnotesize
  \centering
    \begin{tabular}{|c|c|c|c|c|}\hline
    \multirow{2}[4]{*}{Sequence} & \multicolumn{4}{c|}{Average Success Rate(\%)} \\\cline{2-5}
          & SSVM  & SML & SMT & SMM \\\hline
    barbapapa & 94.1176 & \best{94.4356} & 94.2766 & \best{94.4356} \\
    comic & 98.1250 & 98.5417 & \second{98.6458} & \best{98.8542} \\
    map   & \best{98.7603} & 98.6226 & \best{98.7603} & \best{98.7603} \\
    paper & 82.7807 & 86.2032 & \second{87.3797} & \best{88.2353} \\
    phone & 96.6711 & 97.2037 & \second{97.6032} & \best{98.4021} \\
    chart & 53.0337 & \second{62.0225} & 61.1236 & \best{77.5281} \\
    keyboard & 62.3549 & 73.6318 & \second{76.6169} & \best{94.5274} \\
    food  & 85.7585 & 88.0805 & \second{99.3808} & \best{99.6904} \\
    book  & 55.8219 & 71.5753 & \second{74.8288} & \best{81.6781} \\\hline
    \end{tabular}%
    \caption{\footnotesize{Evaluation of our individual components in the average success rate (higher is better). The best result on each sequence is shown in bold font. We find that both metric learning and multi-task learning based approach obtain a higher success rate than the structured SVM approach, and our joint learning approach achieves the best performance.}}
    \label{tab:ours}
\end{table}%

\begin{figure}[t]
\centering
\begin{minipage}[t]{0.49\linewidth}
    \centering
    \includegraphics[width=1\textwidth]{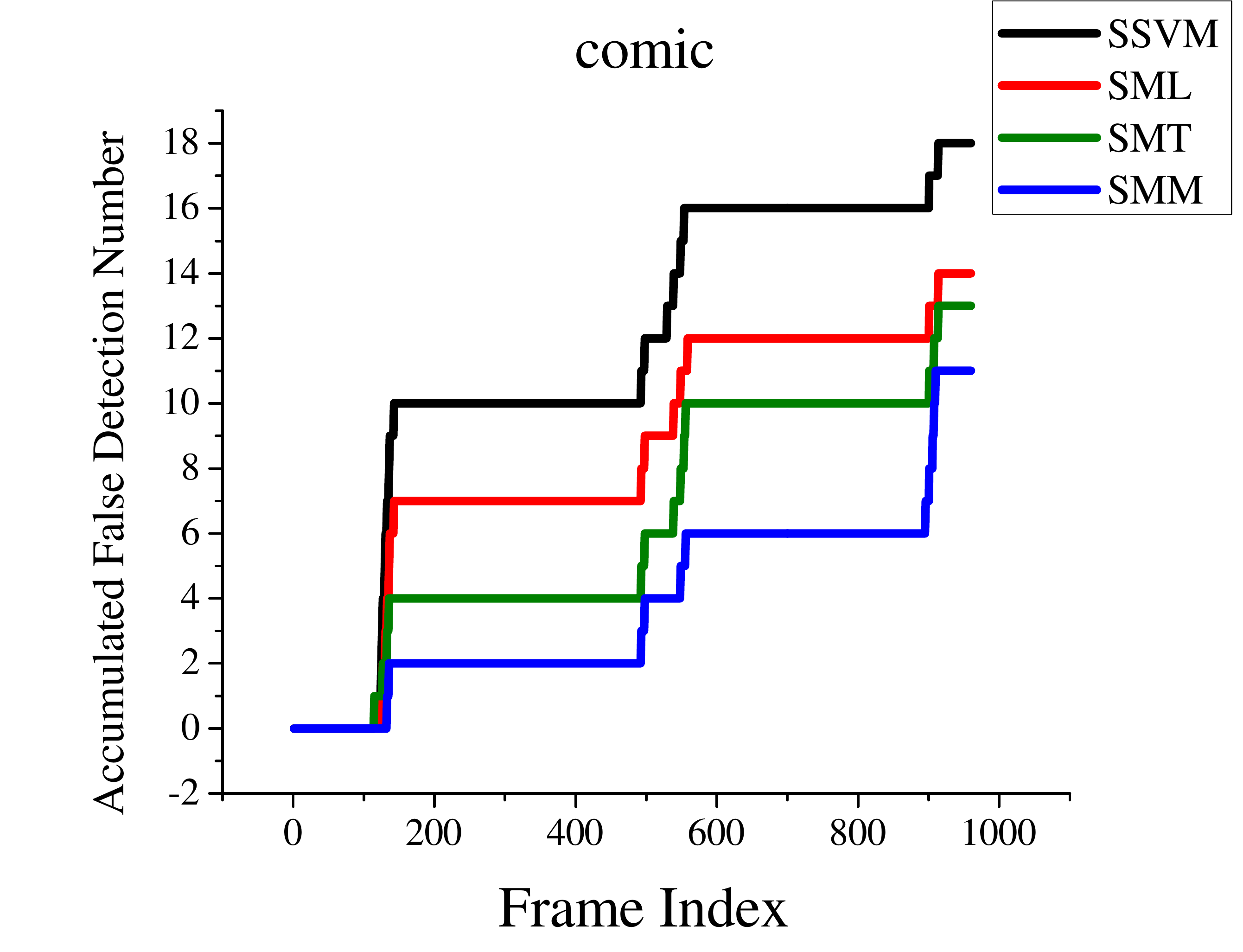}
\end{minipage}
\begin{minipage}[t]{0.49\linewidth}
    \centering
    \includegraphics[width=1\textwidth]{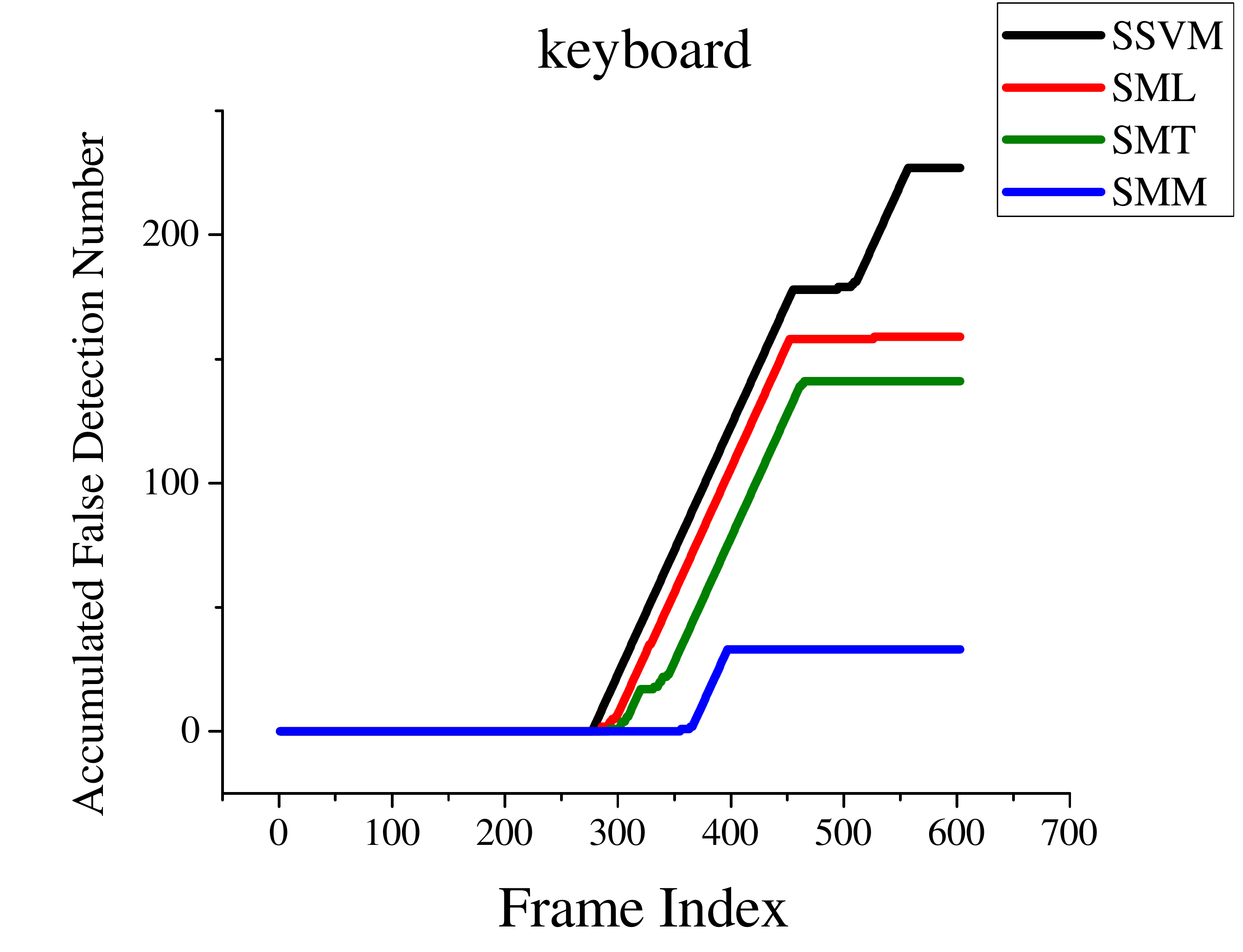}
\end{minipage}
\\
\begin{minipage}[t]{0.49\linewidth}
    \centering
    \includegraphics[width=1\textwidth]{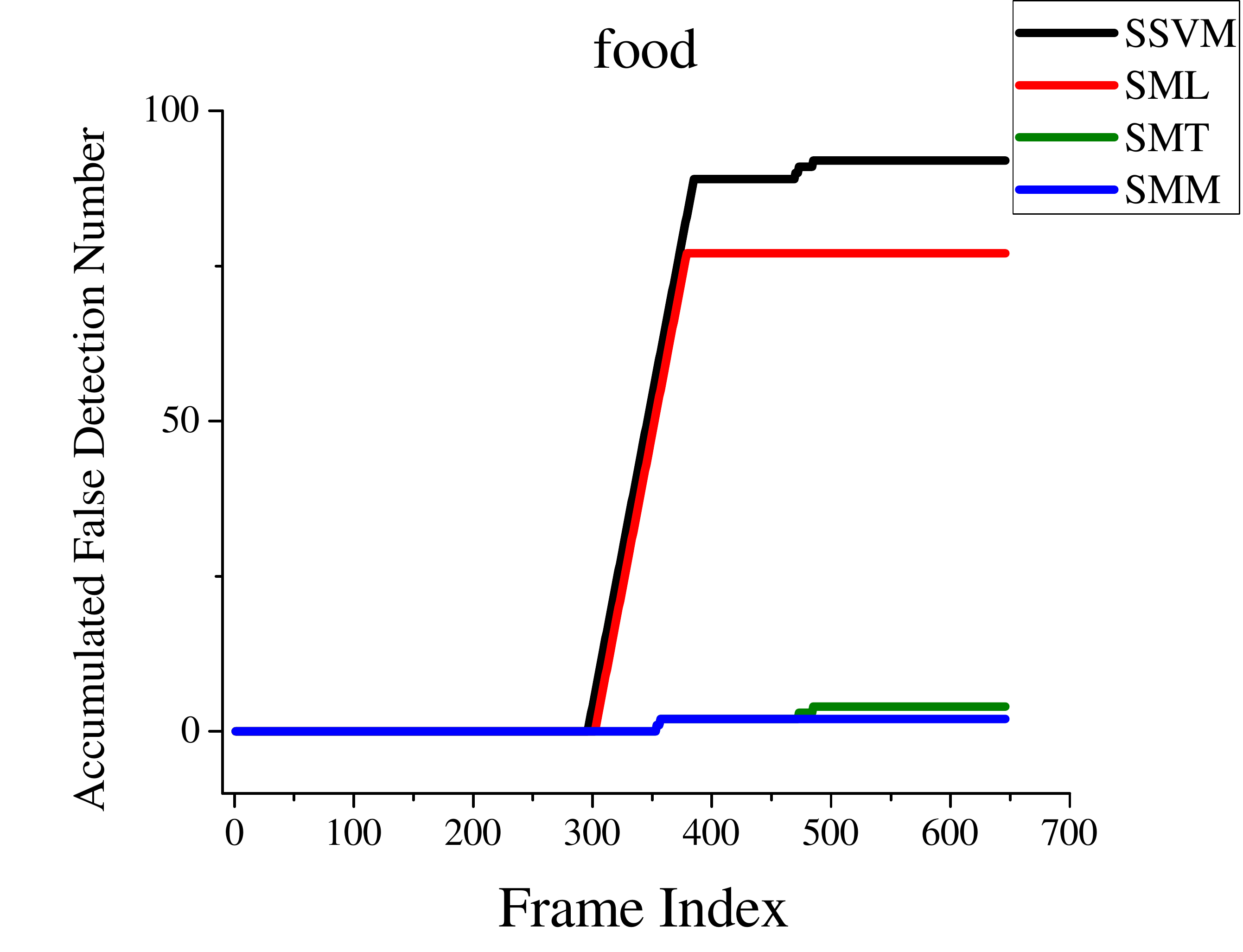}
\end{minipage}
\begin{minipage}[t]{0.49\linewidth}
    \centering
    \includegraphics[width=1\textwidth]{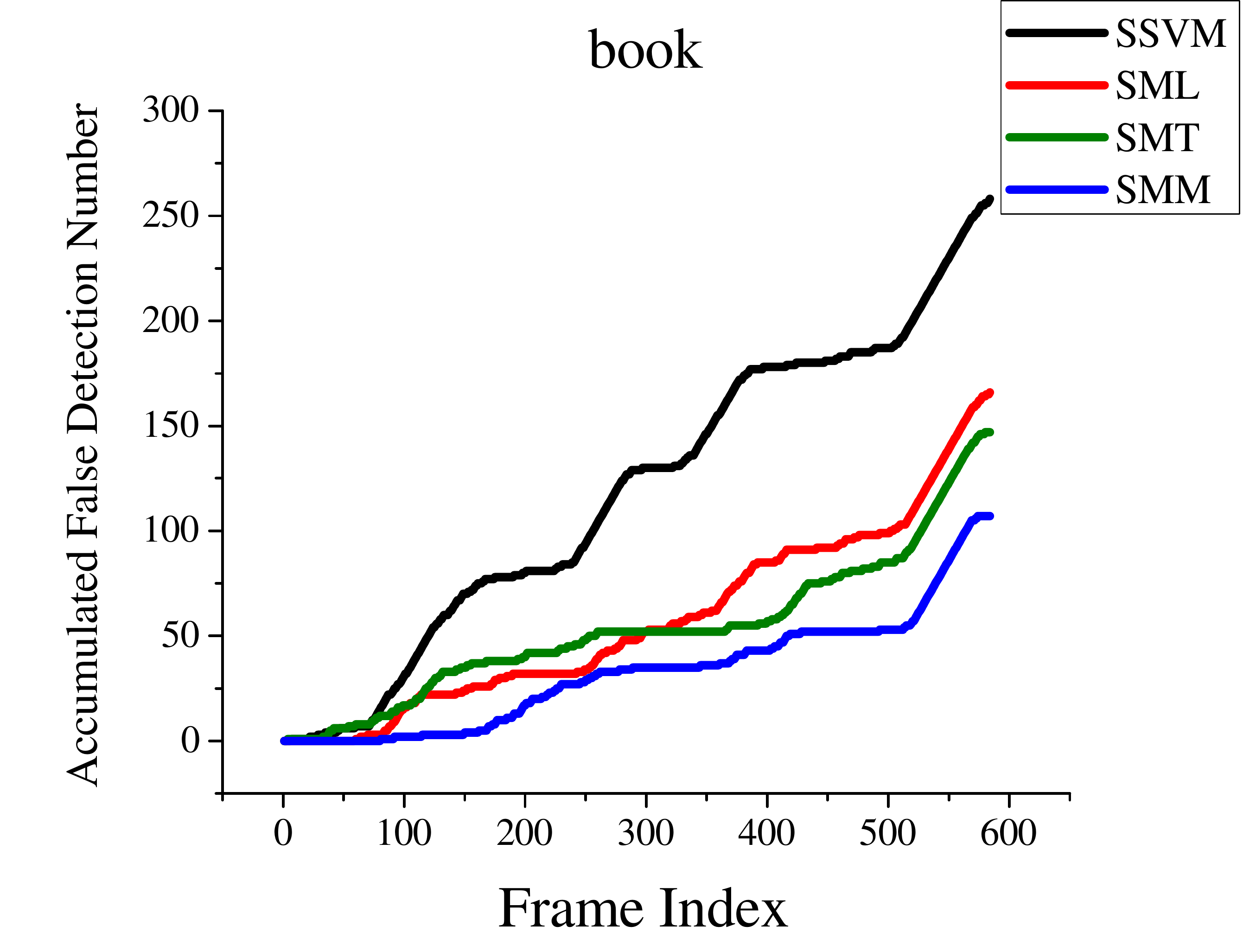}
\end{minipage}
\caption{\footnotesize{Evaluation of our individual components in the accumulated number of falsely detected frames (lower is better). We observe that both metric learning and multi-task learning can improve the robustness of the tracker.}}
\label{fig:indiv}
\end{figure}

From Table~\ref{tab:ours}, we find that the geometric verification based structured learning approach achieves good tracking results in most situations. Furthermore, we observe from Figure~\ref{fig:indiv} that multi-task structured learning guides the tracker to produce a stable tracking result in the complicated scenarios, and metric learning enhances the capability of the tracker to separate keypoints from background clutter. Our approach consisting of all these components then generates a robust tracker.

\section{Conclusion}
In this paper, we have presented a novel and robust keypoint tracker by solving a multi-task structured output optimization problem driven by metric learning. Our joint learning approach have simultaneously considered spatial model consistency, temporal model coherence, and discriminative feature construction during the tracking process.

We have shown in extensive experiments that geometric verification based structured learning has modeled the spatial model consistency to generate a robust tracker in most scenarios, multi-task structured learning has characterized the temporal model coherence to produce stable tracking results even in complicated scenarios with drastic changes, and metric learning has enabled the discriminative feature construction to enhance the discriminative power of the tracker. We have created a new benchmark video dataset consisting of challenging video sequences, and experimental results performed on the dataset have shown that our tracker outperforms the other state-of-the-art trackers.

\section{Acknowledgments}
All correspondence should be addressed to Prof. Xi Li. 
This work is in part supported  by the National Natural Science Foundation of China (Grant No. 61472353), 
National Basic Research Program of China (2012CB316400), 
NSFC (61472353), 863 program (2012AA012505), 
China Knowledge Centre for Engineering Sciences and Technology and the Fundamental Research Funds for the Central Universities.

\fontsize{9.5pt}{10.5pt} \selectfont
\bibliography{zhao}
\bibliographystyle{aaai}

\end{document}